\documentclass[letterpaper, 10 pt, conference]{ieeeconf}  %

\IEEEoverridecommandlockouts                              %
\usepackage{graphicx} 
\usepackage{amsmath} 
\usepackage{amssymb}
\usepackage{comment}
\usepackage{multirow}
\usepackage{color}
\usepackage{url}
\usepackage{subcaption}

\newcommand{\Dtr}{\mathcal{D}_{\text {tr}}}
\newcommand{\Dval}{\mathcal{D}_{\text {val}}}

\newcommand{\So}{\mathcal{S}_{\text {o}}}
\newcommand{\bsy}{q}
\newcommand{\bsystems}{b}

\newcommand{\nsamp}{N}
\newcommand{\norm}[1]{\left\lVert#1\right\rVert}
\newcommand{\free}{\mathcal{T}}

\newcommand{\regpar}{\gamma}

\title{\LARGE \bf
Synthetic data generation for system identification: leveraging knowledge transfer from similar systems
}

\author{Dario Piga$^{1}$, Matteo Rufolo$^{1}$, Gabriele Maroni$^{1}$, Manas Mejari$^{1}$, Marco Forgione$^{1}$
\thanks{Corresponding author: {\tt dario.piga@supsi.ch}.}
\thanks{This project was partially supported by the   Eurostars programme, project ``\emph{E3093 - SLIMPEC: A Software suite for LearnIng-based embedded
Model PrEdictive Control}''. The activities of G. Maroni are partially supported  by HASLER STIFTUNG under the project ``\emph{Accurate and Interpretable Deep Learning models for Peak Load Forecasting}'', project number: 23040.}
\thanks{$^{1}$SUPSI-DTI-IDSIA, Dalle Molle Institute for Artificial Intelligence, Lugano, Switzerland.}}

\begin{document}

\maketitle
\thispagestyle{empty}
\pagestyle{empty}

\begin{abstract}
This paper addresses the challenge of overfitting in the learning of dynamical systems by introducing a novel approach for the generation of synthetic data, aimed at enhancing model generalization and robustness in scenarios characterized by data scarcity. Central to the proposed methodology is the concept of knowledge transfer from systems within the same class. Specifically, synthetic data is generated through a pre-trained meta-model that describes a broad class of systems to  which the system of interest is assumed to belong. Training data serves a dual purpose: firstly, as input to the pre-trained meta model to discern the system's dynamics, enabling the prediction of its behavior and thereby generating synthetic output sequences for new input sequences; secondly, in conjunction with synthetic data, to define the loss function used for model estimation. A validation dataset is used to tune a scalar hyper-parameter balancing the relative  importance of training and synthetic data in the definition of the loss function. The same validation set can be also used for other purposes, such as early stopping during the training, fundamental  to avoid overfitting  in case of small-size training datasets. 
The efficacy of the approach is shown through a numerical example that highlights the advantages of integrating synthetic data into the system identification process. 
\end{abstract}

\section{Introduction}
The performance of system identification algorithms, as well as other machine learning tools, is greatly dependent on the amount and quality of the training data. This dependence becomes challenging in situations where data is scarce and expensive to acquire, leading to overfitting  when complex models are adopted.

In  the machine learning literature, two main strategies  have been developed to address this challenge: data augmentation and the use of synthetic data \cite{shorten2019survey,nikolenko2021synthetic}, each with its own   benefits. Data augmentation is a widely used technique in various domains, such as image and natural language processing. It involves modifying existing data samples to create new variations. This technique helps to enrich the dataset and introduce diversity, which improves the model's ability to generalize from a limited number of samples. However, data augmentation is limited by the scope and characteristics of the original data. 
On the other hand, the production and application of synthetic data offers a broader solution. Synthetic data goes beyond simply modifying existing data. It entails generating completely new, artificial datasets that mimic the statistical characteristics of real-world data. This approach not only overcomes the constraints of data augmentation but also provides a method to generate large and diverse datasets. However, generating reliable synthetic data might be challenging and may not be always possible. 

To the best of our knowledge, only few contributions are available for data augmentation/synthetic data generation in system identification. The work \cite{formentin2019nonlinear} addresses identification of dynamical systems described in terms of nonlinear finite impulse response, where synthetic regressors (with no corresponding outputs) are created  by slightly perturbing  the original regressors, and manifold regularization is applied using these new synthetic regressors.  The contribution~\cite{wakita2023data} proposes  data augmentation by modifying the original time sequences through jittering and slicing, with application to estimation of dynamical model describing  harbor manouvers in maritime autonomous surface ships.  

The popular physics-informed deep learning paradigm, introduced by Raissi \emph{et al.} in \cite{raissi2019physics}, can be seen as a method for working with synthetic datasets. Indeed, physical laws are integrated into the training procedure to restrict the set of admissible solutions or to enforce known dynamics, thus enhancing model robustness, generalization, and explainability, even in the case of small-sized training datasets. Essentially, a regularization term is added to the fitting loss of the available training data. This regularization is formed by considering collocation points (distributed in the spatio-temporal domain), where the available \emph{Partial Differential Equation} (PDE) describing the system should be satisfied.

In this paper, we leverage the power of synthetic data in scenarios of small-size training data. We exploit the new modeling paradigm recently proposed by some of the authors in \cite{forgione2023context} in order to derive  a \emph{meta-model}  (describing a broad class to which the query system\footnote{We refer to the system to be identified as a ``query system'', the data generated from which is used as a context for a meta-model.} is assumed to belong) that is used to generate a set of synthetic data. As discussed in \cite{forgione2023context}, such a meta-model is trained on a potentially infinite stream of synthetic data, generated by simulators with randomly generated settings. The proposed approach harnesses the power of Transformers, commonly used for Natural Language Processing~\cite{vaswani2017attention} and representing the key technology behind Large Language Models \cite{achiam2023gpt, touvron2023llama}. At the inference time, a (typically short) input-output sequence generated by the query system (namely, available training data) is used as a context for the Transformer, which implicitly discerns the dynamics of the system, enabling predictions of its behavior. By querying the meta-model with different new input sequences, a potentially infinite stream of synthetic data  is then constructed. From a different perspective, since the available training data are used as context to the meta model to generate synthetic output samples, the technique can also be seen as a data augmentation strategy, as the training data are manipulated to form new data. Overall, synthetic data for the query system is generated   leveraging knowledge transfer from systems within the same class, all integrated within the pre-trained Transformer.  Once the synthetic input/output sequences become available, a model (either gray or black box) of the query system, fitting both the training and the synthetic data, is estimated using any system identification tool.

The paper is organized as follows: The problem addressed is described in detail in Section \ref{Sec:probdesc}. The generation of synthetic data is discussed in Section \ref{sec:data_gen}, which includes a description of the encoder-decoder Transformer architecture used for synthetic data generation and describing the class of systems. Parametric system identification with synthetic data is described in Section \ref{sec:learning}. A numerical example is reported in Section \ref{Sec:example} to demonstrate the effectiveness of the approach and to show the advantages of using synthetic data in identification problems with small datasets. Concluding remarks and directions for future research are presented in Section \ref{sec:conclusions}.

\section{Problem description} \label{Sec:probdesc}
We consider a dataset $\Dtr$ containing a sequence of input-output pairs generated by a dynamical system $\So$, denoted as $\Dtr=\left\{(u_t,y_t)\right\}_{t=1}^T$, where $u_t$ and $y_t$ respectively represent the input and output of $\mathcal{S}_{\text o}$ at time step $t$, with $t=1,\ldots,T$. For simplicity, and without loss of generality, we assume that the system $\mathcal{S}_o$ is single-input single-output (SISO), i.e., $u_t, y_t \in \mathbb{R}$.

Our focus is on a standard system identification problem, aiming to fit a parametric causal dynamical model $\mathcal{M}(\cdot;\theta)$ to the training dataset $\Dtr$. Here, $\theta$ denotes the parameters describing the model $\mathcal{M}(\cdot;\theta)$, which maps an input sequence $u_{1:t}$ up to time $t$ to an output $\hat{y}_t$, as defined by:
\begin{equation}
\hat{y}_t(\theta) = \mathcal{M}(u_{1:t};\theta),
\end{equation}
where the dependence of $\hat{y}_t(\theta)$ on $u_{1:t}$ is omitted to simplify the notation.

Additionally, our goal is to augment the training dataset $\Dtr$ by generating a potentially infinite-dimensional set of synthetic input-output trajectories $\left\{\tilde{u}^{(i)}_t, \tilde{y}^{(i)}_t\right\}_{t=1}^{\tilde{T}}$, where the superscript $i=1,2,\ldots$ is used to denote the $i$-th synthetic trajectory, and $\tilde{T}$ is the length of the synthetic trajectories, which are all assumed to  have the same length just to keep the notation simple. The  trajectories $\tilde{u}^{(i)}_{1:t}, \tilde{y}^{(i)}_{1:t}$ are assumed to be drawn from a probability distribution $P(\tilde{u},\tilde{y})$ which ideally should be the same underlying distribution generating the training dataset $\Dtr$.

In the following section we show how to generate the synthetic output sequence $ \tilde{y}^{(i)}$ for given input trajectories $\tilde{u}^{(i)}$.

\section{Synthetic data generation} \label{sec:data_gen}
In order to generate the synthetic input-output sequences $\tilde{u}^{(i)}, \tilde{y}^{(i)}$, we assume that: (i) the system $\So$ we aim to model belongs to a broad class of dynamical systems; (ii) the meta-model, which describes the behavior of this class and is used to generate synthetic data, has been pre-trained using data from systems within the class.  This approach is based on the system class modeling paradigm developed by some of the authors in \cite{forgione2023context}, which is briefly reviewed in the  
following  for self-consistency of the paper.

\begin{figure*}[!bt]
\centering
\includegraphics[width=.95\textwidth]{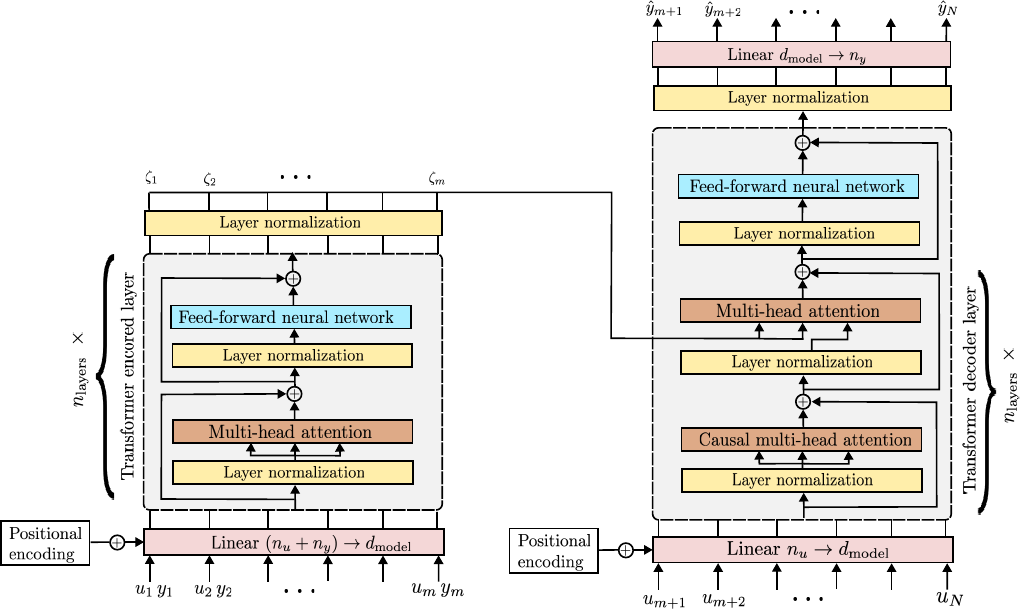}
\caption{Encoder-decoder Transformer model for the class of systems, used to generate synthetic data. The Transformer is  characterized by: number of layers ($n_{\text{layers}}$), model dimensionality per layer ($d_{\text{model}}$), number of attention heads ($n_{\text{heads}}$), and context window length ($m$).}
\label{fig:encoder_decoder_arch}
\end{figure*}

A Transformer with an encoder-decoder  architecture, illustrated in Fig.~\ref{fig:encoder_decoder_arch}, is used   for generating  synthetic data of the system $\mathcal{S}_{\text o}$. It is basically the standard Transformer architecture with attention mechanism proposed in~\cite{vaswani2017attention},  adapted in \cite{forgione2023context} to process sequences of real input/output data, and further modified in \cite{piga2023adaptation} with positional encoding used instead of positional embedding.

The encoder processes an input/output sequence $u_{1:m}, y_{1:m}$ (typically referred to as `context') and generates an embedding sequence $\zeta_{1:m}$, which is   processed by the decoder along with a test input $u_{m+1:\nsamp}$ (the latter subject to causal restriction) to produce the sequence of predictions $\hat{y}_{m+1:\nsamp}$ up to step $\nsamp$. The parameters $\phi$ of the Transformer are obtained by randomly sampling systems from the class, and then minimizing over $\phi$ the empirical loss according to a supervised learning paradigm:
\begin{multline}
\label{eq:simulation_objective_samples}
J =
\frac{1}{\bsystems}
\sum_{i=1}^{\bsystems}
\norm{y_{m+1:\nsamp}^{(i)} - \free_\phi (u_{1:m}^{(i)}, y_{1:m}^{(i)}, u_{m+1:\nsamp}^{(i)})}^2
,
\end{multline}
where: $\free_\phi (u_{1:m}^{(i)}, y_{1:m}^{(i)}, u_{m+1:\nsamp}^{(i)})$ denotes the output $\hat{y}_{m+1:\nsamp}$ of the Transformer when fed with a context $\{ u_{1:m}^{(i)}, y_{1:m}^{(i)}\}$ and query test input  $u_{m+1:\nsamp}^{(i)}$; and $\bsystems$ denotes the number of randomly generated systems, each providing an input/output sequence $\{u^{(i)}_{1:\nsamp}, y^{(i)}_{1:\nsamp}\}$, split to form the context and to create the empirical loss $J$ in \eqref{eq:simulation_objective_samples}.  
\emph{Mini-batch gradient descent} is then used to minimize $J$,  with  $\bsystems$ new systems and sequences resampled \emph{at each iteration}. 
It should be noted that, in practical scenarios, simulators can be employed to produce the input/output sequences $\{u^{(i)}_{1:\nsamp}, y^{(i)}_{1:\nsamp}\}$ for pre-training the Transformer. Consequently, it becomes possible to create a diverse range of datasets for estimating the Transformer's parameters by adjusting software configurations (for instance, physical parameters) based on insights from the application domain.

The Transformer, pre-trained on data simulated from systems randomly selected from a particular class, serves as an extensive meta-model for that class. It gains the ability to infer the behavior of a specific query system $\So$ directly (specifically, through zero-shot in-context learning) from the existing training dataset $\Dtr$, which implicitly contains the main  characteristics of the system $\So$. The training dataset provides context for the pre-trained Transformer, enabling it to generate synthetic outputs $\tilde{y}_{1:\tilde{T}}$ for any query input sequence $\tilde{u}_{1:\tilde{T}}$. Consequently, this approach allows  the generation of a potentially infinite stream of synthetic input/output data, by implicitly leveraging knowledge transfer  from systems within the same class.

\section{Learning with augmented data} \label{sec:learning}

The  parameters $\theta$ characterizing the model  $\mathcal{M}(\cdot;\theta)$ of the query system $\So$  are then estimated by minimizing the following expected loss, which considers both actual training data (from$\So$) and artificially generated synthetic data (from the pre-trained Transformer):

\begin{align}  
\hat{\theta} =  \arg\min_{\theta \in \Theta}  & \frac{1}{T}\sum_{t=1}^T \ell\left(y_t, \hat{y}_t(\theta) \right) + \nonumber \\
& + \regpar \mathbb{E}_{P(\tilde{u},\tilde{y}}) \left[ \frac{1}{\tilde{T}} \sum_{t=1}^{\tilde{T}}\ell\left(\tilde{y}_t, \hat{\tilde{y}}_t(\theta)\right)\right]    \label{eqn:lossE}
\end{align}
where $\hat{\tilde{y}}_t(\theta)$ represents the model's output at time $t$ for a synthetic input sequence $\tilde{u}_{1:t}$, and $\ell$ denotes a chosen loss function, such as the squared error defined by:
\begin{align} \label{eqn:L}
\ell\left(y_t, \hat{y}_t(\theta) \right) = \left( y_t - \hat{y}_t(\theta)\right)^2.
\end{align}

The term $\regpar$ is a non-negative regularization hyperparameter that balances the influence of training and synthetic data in the model fitting process. A value of $\regpar=0$ focuses the model on fitting the training data alone, potentially leading to overfitting, especially with overly complex models or small datasets. Increasing $\regpar$ elevates the significance of synthetic data, which might introduce biases from the synthetic data generation process. 
The choice of $\regpar$ should ideally be determined by the quality and amount of real training data and the reliability of synthetic data. 
In this paper, $\regpar$ is selected through hold-out validation, using a portion of the training dataset $\Dtr$ or a separate validation dataset $\Dval$ to regulate the balance between real and synthetic data influences. 
 As discussed in the example in Section \ref{Sec:example}, we employ early stopping criteria to prevent overfitting, especially when relying on a small training dataset and when synthetic data is not utilized. Consequently, the same validation dataset used for early stopping can also be adopted to adjust $\regpar$, ensuring that no additional data is required beyond what is already used for model estimation with early stopping.

The expected value in \eqref{eqn:lossE} is  approximated by applying   $\bsy$ synthetic    sequences $\tilde{u}^{(i)}$ (with $i=1,\ldots,\bsy$) as input of the decoder and then generating corresponding  output sequences $ \tilde{y}^{(i)}$. The estimation problem \eqref{eqn:lossE}  thus becomes: 

\begin{align}  
\hat{\theta} =  \arg\min_{\theta \in \Theta}  & \frac{1}{T}\sum_{t=1}^T \ell\left(y_t, \hat{y}_t(\theta) \right) + \nonumber \\
& + \regpar \frac{1}{\bsy} \frac{1}{\tilde T} \sum_{i=1}^{\bsy}  \sum_{t=1}^ {\tilde T}   \ell\left(\tilde{y}^{(i)}_t,\hat{\tilde y}^{(i)}_t(\theta)\right). \label{eqn:loss_sample}
\end{align}

The optimization problem \eqref{eqn:loss_sample} is then solved through mini-batch stochastic gradient descent by generating, at each iteration, $\bsy$ new synthetic input/output sequences.

\section{Example} \label{Sec:example}

In this section we illustrate the effectiveness of the proposed method with a numerical example  focused on the identification of Wiener-Hammerstein models, which   provide  a block-oriented representation for numerous nonlinear dynamical systems encountered in practice~\cite{giri2010block}. The pre-trained meta model, along with the Python scripts for synthetic data generation, are available in the GitHub repository associated with this  paper \cite{gitSDGPi24}.

\subsection{Data-generating system}
\begin{figure}[!bt]
    \centering
    \includegraphics[height=20pt]{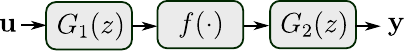}
    \caption{The Wiener-Hammerstein system structure.}
    \label{fig:WHsystem}
\end{figure}

For data generation, we consider a SISO stable Wiener-Hammerstein dynamical system $\So$, as visualized in Fig.~\ref{fig:WHsystem}, with the structure $G_1$--$F$--$G_2$, where $G_1$ and $G_2$ are Linear-Time-Invariant (LTI) blocks, and $F$ is a static non-linearity sandwiched between  $G_1$ and $G_2$. The LTI blocks
$G_1$ and $G_2$ are randomly chosen, with dynamical   order   between $1$ to $10$, and poles  randomly generated and constrained to have  magnitude  and phase    in the range $(0.5,0.97)$ and $(0, \pi/2)$, respectively.  The   non-linear block $F$ is   a feed-forward neural network with one hidden layer of size  $32$,  and parameters randomly drawn   from a  Gaussian distribution. 

A training dataset $\Dtr$ and a validation dataset $\Dval$, of lengths $T=250$ and $T_{\text{val}}=100$ respectively, are generated by exciting the system with a white input signal drawn from a normal distribution with zero mean and unit variance. The output samples are corrupted by white Gaussian noise with a standard deviation of $\sigma_e=0.35$, corresponding to a Signal-to-Noise Ratio (SNR) of $9.1$ dB. 

A test dataset $D_{\text{test}}$ consisting of $4,000$ samples is also generated. For simplification in evaluating the model's performance, the test outputs are considered without noise. It is important to note that, in practical applications, the size of the test dataset should ideally not be $16$ times larger than that of the training dataset. Nevertheless, employing a large test dataset enhances the statistical significance of the results obtained. To further increase the statistical reliability of these results, a Monte Carlo study comprising $100$ runs is conducted, with new data-generating system $\So$, training, validation, and test datasets being generated for each run.

\subsection{Synthetic Data Generation}
 The Transformer $\free_\phi$ describing the considered class of systems was pre-trained as described in  \cite{forgione2023context} using an Nvidia RTX 3090 GPU. 
The Transformer, which comprises $5.6$ million weights, is characterized by $n_{\text{layers}}=12$ layers, $d_{\text{model}}=128$ units in each layer, $n_{\text{heads}}=4$ attention heads, and an encoder's context window length of $m = T = 400$. Such a pre-trained Transformer describes the class of SISO Wiener-Hammerstein systems with LTI blocks of dynamical order up to $10$.

The context provided to the encoder is the input-output training sequence. Synthetic output samples are generated by applying query input sequences $\tilde{u}_{1:\tilde{T}}$ of length $\tilde{T}=200$, which share the same statistical distribution as the training input.

\subsection{Parametric model}
We aim at estimating a parametric model $\mathcal{M}(\cdot;\theta)$ describing the behavior of the query Wiener-Hammerstein data-generating system $\So$.  

As a structure for the parametric model  $\mathcal{M}(\cdot;\theta)$  we consider a Wiener-Hammerstein one, where the LTI blocks have a dynamical order of $10$. The non-linear block $F$ is a feed-forward neural network with one hidden layer of size $32$.   Overall, the considered model structure is characterized by $137$ parameters. Such a parametric model $\mathcal{M}$ have  the same structure of the class of systems described by the  pre-trained Transformer $\free_{\phi}$. Therefore, the prior information about the class to which the query system is supposed to belong has been used to both generate synthetic data and select the structure of the parametric model $\mathcal{M}$.

The loss \eqref{eqn:loss_sample} is minimized through stochastic gradient descent for  maximum  $6000$ iterations, with batch size $q=1$, and by generating at each iteration new synthetic input-output sequences. In minimizing the loss, we exploited the approach in~\cite{forgione2021dynonet} for fast differentiation of the linear dynamical blocks. An early stopping criterion~\cite{girosi1995regularization} is also adopted to avoid overfitting, which mainly occurred when no synthetic data was used (i.e., for hyperparameter $\gamma=0$) or for small values of $\gamma$ (i.e., $\gamma \leq 1$).    

The hyperparameter $\gamma$ is selected through a coarse grid search, considering the following  values: $0, 0.1, 1, 10, 20, 30,  50, 100, 200$ and by minimizing the mean squared error (MSE) over the validation dataset $\Dval$, which  is defined as:

\begin{align}
\text{MSE} = \frac{1}{T_{x}} \sum_{t=1}^{T_{x}} \left( y_t - \hat{y}_t(\theta)\right)^2,
\end{align}
where $T_x$ is the length of the dataset where the MSE is computed ($T_x = T_{\text{val}}=100$ for validation set), $y_t$ denotes the true output at time step $t$, and $\hat{y}_t(\theta)$ indicates the  predicted model's output. We remark that the same validation dataset is adopted both for early stopping and to select the hyperparameter $\gamma$.

\subsection{Results}

\begin{figure*}[!t]
    \centering
    \begin{subfigure}[b]{0.48\textwidth}
        \includegraphics[width=\textwidth]{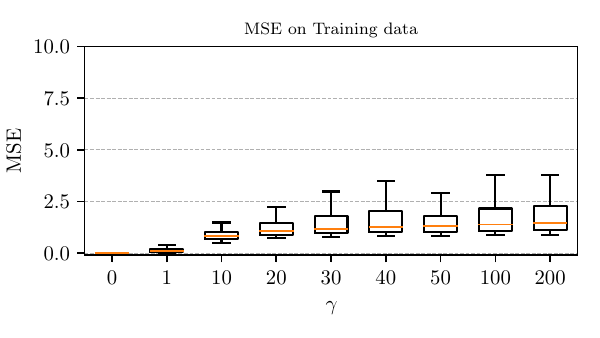}
        \caption{MSE \emph{vs} $\gamma$ on training data.}
        \label{fig:MSE_tr}
    \end{subfigure}
    \hfill
    \begin{subfigure}[b]{0.48\textwidth}
        \includegraphics[width=\textwidth]{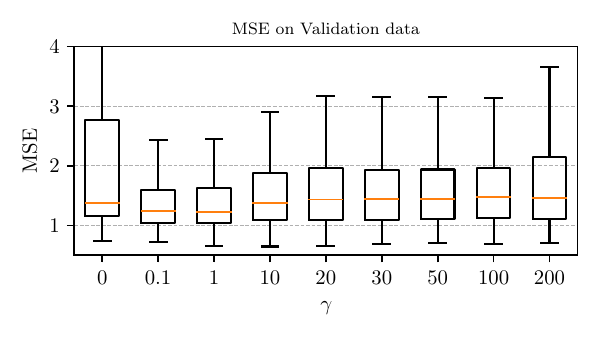}
        \caption{MSE \emph{vs} $\gamma$ on validation data.}
        \label{fig:MSE_val}
    \end{subfigure}
    \caption{Impact of regularization hyparameter $\gamma$  on mean squared error in training (left panel) and validation (right panel) dataset. Boxplots of average squared error   over $100$ Monte Carlo runs. In the right panel, the vertical limit is set   to $4$ for a better visualization of the boxplots associated to $\gamma \neq 0$.}
    \label{fig:MSE}
\end{figure*}

\begin{figure}[!b]
\centering
\includegraphics[width=\linewidth]{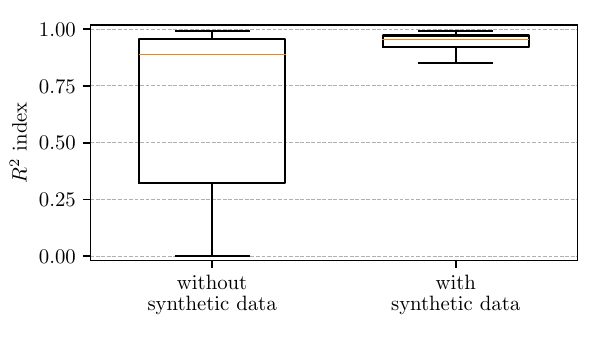}
\caption{Impact of synthetic data on $R^2$ performance in test dataset. Boxplots of $R^2$ coefficients over $100$ Monte Carlo runs: without using synthetic data (left);  using synthetic data (right).}
\label{fig:r2index}
\end{figure}

Fig.~\ref{fig:MSE} shows the boxplots of the Mean Squared Error  on the training (left panel) and validation dataset (right panel) obtained for different values of the hyperparameter $\gamma$, with   $\gamma=0$ corresponding  to fit only the available training data. To illustrate the presence of overfitting, the MSE on training data is not the one obtained by the model estimated with early-stopping, but the one achieved when the maximum number of iterations is reached.  Results in the figure show the regularization role played by synthetic data. Indeed, for $\gamma=0$ the MSE in the validation data is about $5$  times larger than the one in training, while similar MSEs are achieved in training and validation when the relevance of  synthetic data w.r.t. training data is significant (i.e., for $\gamma \geq 10$). We also notice a benefit in validation performance when synthetic data is used ($\gamma >0$). Improvement is  visible, although we observe that when  the ratio of synthetic data in the loss is more than 10 times larger than training data, performance in validation  decreases, intuitively  due to the fact that measured training data, although affected by noise, are more reliable than synthetic data generated by the meta-model, which can be affected by epistemic uncertainty.

Fig.~\ref{fig:r2index} shows results obtained on the test dataset $\mathcal{D}_{\text{test}}$. For a better interpretation of the performance, the $R^2$ coefficient is plotted in the figure instead of the MSE, where the $R^2$ is defined as:
\begin{align}
R^2 = 1 - \frac{\sum_{t=1}^{T_{x}} \left( y_t - \hat{y}_t(\theta)\right)^2}{\sum_{t=1}^{T_{x}} \left( y_t - \bar{y}\right)^2}
\end{align}
where $\bar{y}$ is the mean output  and $T_x=4,000$ is the length of the test dataset. In the test, we only considered (and thus reported) a comparison between performance achieved without using synthetic data and performance achieved using synthetic data, with regularization hyperparameter $\gamma$ selected, at each Monte Carlo run, through hold-out validation. As expected from the results in validation, also test results show a significant improvement in the model performance thanks to the usage of synthetic data, with a median of the $R^2$ coefficient which improved from $0.889$ (in case of no synthetic data) to $0.956$.

\section{Conclusions}  \label{sec:conclusions}

In this paper, we demonstrate the feasibility of generating an extensive stream of synthetic data for system identification by employing knowledge transfer from analogous systems. This allows to mitigate the challenges posed by data scarcity, offering  enhancements in model performance and generalization capabilities. A practical numerical example highlights the efficacy of the methodology, showcasing an   increasing of the $R^2$ coefficient when synthetic data is integrated with traditional training datasets.

Current research directions are focused on:
\begin{itemize}
    \item Refinement and scaling-up of the meta-model  to encompass a broader spectrum of dynamical systems. This  ensures its applicability across a diverse range of system identification scenarios.
\item Estimation of the uncertainty associated with the meta-model’s outputs. This allows to reformulate the minimization of the loss with synthetic data as a Maximum Likelihood estimation problem. This advancement will allow for a proper weighting of synthetic data based on their reliability, ensuring that less certain synthetic samples exert a smaller influence on the model estimation process compared to more reliable ones and actual training data.
\item Enhancing Bayesian estimation algorithms, such as Gaussian Process Regression, by incorporating the meta-model's output as a prior. This strategy is particularly beneficial in areas of the input space lacking observations, where the model increasingly relies on the prior. This incorporation seeks to utilize the knowledge derived from analogous systems,  encapsulated by the meta-model, to make more informed inferences in domains with few observations.
\end{itemize}

\addtolength{\textheight}{-12cm}   


\bibliographystyle{IEEEtran} 
\bibliography{root}


\end{document}